\documentclass[runningheads]{llncs}
\usepackage[square, numbers]{natbib}
\usepackage{graphicx, caption}
\usepackage{subfig}
\usepackage{multirow}
\usepackage{makecell}
\usepackage{booktabs}
\usepackage[english]{babel}
\usepackage[autostyle]{csquotes}

\begin{document}
\title{Improving Cause-of-Death Classification from Verbal Autopsy Reports}
\titlerunning{Improving Cause-of-Death Classification from Verbal Autopsy Reports}
%
\author{Thokozile Manaka\inst{1}\orcidID{0000-0001-9910-4480} \and
Terence van Zyl\inst{2}\orcidID{0000-0003-4281-630X} \and
Deepak Kar\inst{3}\orcidID{0000-0002-4238-9822}}

\authorrunning{T. Manaka et al.}
%
\institute{School of Computer Science and Applied Mathematics, University of The Witwatersrand, Johannesburg\\ 
\email{thokozilemanaka@wits.ac.za} \and
Institute for Intelligent Systems, University of Johannesburg, Johannesburg\\
\email{tvanzyl@uj.ac.za} \and
School of Physics, University of The Witwatersrand, Johannesburg\\
\email{deepak.kar@wits.ac.za}}

%
\maketitle              
\begin{abstract}
In many lower-and-middle income countries including South Africa, data access in health facilities is restricted due to patient privacy and confidentiality policies. Further, since clinical data is unique to individual institutions and laboratories, there are insufficient data annotation standards and conventions. As a result of the scarcity of textual data, natural language processing (NLP) techniques have fared poorly in the health sector. A cause of death (COD) is often determined by a verbal autopsy (VA) report in places without reliable death registration systems. A non-clinician field worker does a VA report using a set of standardized questions as a guide to uncover symptoms of a COD. This analysis focuses on the textual part of the VA report as a case study to address the challenge of adapting NLP techniques in the health domain. We present a system that relies on two transfer learning paradigms of monolingual learning and multi-source domain adaptation to improve VA narratives for the target task of the COD classification. We use the Bidirectional Encoder Representations from Transformers (BERT) and Embeddings from Language Models (ELMo) models pre-trained on the general English and health domains to extract features from the VA narratives. Our findings suggest that this transfer learning system improves the COD classification tasks and that the narrative text contains valuable information for figuring out a COD. Our results further show that combining binary VA features and narrative text features learned via this framework boosts the classification task of COD. 
\keywords{Natural Language Processing \and Transfer Learning \and Monolingual Learning \and Multi-domain Adaptation \and Cause of Death}
\end{abstract}
\section{Introduction}
Natural language processing (NLP) methods have played an important role in extracting useful information from unstructured narrative text for classification tasks. Most of these applications have been in the general English domain with convolutional neural networks (CNNs) \citep{Kim, Zhangx, Conneau}, recurrent neural networks (RNNs) \cite{See_2017} and attention based methods \cite{Zlin, Peters}. The medical domain also saw attention based methods used in \cite{Alsentzer, Lee}, RNNs \cite{Jin} and CNNs \cite{Zheng}.

The development of NLP models in the health domain has however shown to be advancing quite slowly in comparison to the general English domain. This is due to a limited access to shared annotated datasets across health institutions and laboratories because of patient privacy and confidentiality policies. There are also insufficient common conventions and standards of annotating clinical data for training and bench-marking NLP applications in this domain \cite{Ohno-Machado}. Sometimes data is not available at all as is the case with many lower-and-middle-income countries where the bulk of fatalities occur outside of medical facilities and no physical autopsies are done \cite{UnitedNations}. The World Health Organization (WHO) has endorsed the use of a verbal autopsies (VAs) in these countries to find the cause of death (COD). These are records of interviews about the events surrounding an uncertified cause of death. As with the language of health reports, this public health report necessitates its own domain-specific development and training as NLP models developed for the general English language text do not generalize well on its narratives.

Transfer learning \cite{kooverjee2022investigating,kooverjee2020inter,karim2020deep,van2020unique,variawa2020rules} allows for knowledge derived from tasks rich in data to be applied to tasks, languages, or domains where data is limited. It consist of two steps, pretraining on one task or domain (source) and domain adaptation where the learned representations are used in a different task, domain or language (target). \citet{Kim} shows that transfer learning has advanced the development of NLP techniques through natural language modeling as a source task and \citet{Pikuliak} reports that the limited data access that results in declined performances of NLP techniques can be addressed with cross-lingual and monolingual learning techniques. 

While domain adaptation assumes that the source domain has abundant training data and aims to use the knowledge learned here to aid the tasks in the target domain which has limited resources, many text classification tasks are domain-dependent in the sense that a text classifier trained on one set of data is likely to under perform on another set as is with the case of labeled and unlabeled or unseen  data. Also, while the most used domain adaptation technique is the one-source-one-target approach, \citet{XChen} shows that multi-domain text classification, where labeled data is present for many domains, but in low amounts for effective training of a text classifier is a more plausible reality. By nature of how it is collected and transcribed, we contend that a VA report represents one actual multi-domain context, and the NLP technologies used on it will thus require a sufficient amount of data from a number of domains. 

Motivated by the multi-domain text classification task where data from a number of domains is fused together for training on a feature or classifier level, we propose a technique that leverages the two transfer learning paradigms of monolingual learning and multi-domain adaptation via the use of embeddings from ELMo pretrained in the English domain and those from BERT pretrained in the biomedical domain. The idea is that the biomedical domain is a subset of the English domain \cite{b21} and therefore by using character-level information from the English domain when computing VA embeddings, grammatical and syntax errors in VA reports will be well taken care of. Additionally, word-level information from the medical domain will capture biomedical relations like symptom interactions across different diseases \cite{Jin}, information that is crucial for COD classification.

This paper's clinical and technical contributions are: 
\begin{enumerate}
    \item A transfer learning approach that improves the cause-of-death (COD) classification task of narrative text features of a verbal autopsy (VA) report by making the most out of understanding the domain adaption process.     
    \item Refined verbal autopsy text representations more suited for the COD classification task.    
    \item We show that multi-domain adaptation for text classification achieves better recall scores than currently used text representation techniques.
\end{enumerate}

\section{Background}

\subsection{Verbal Autopsies}
More than a half of the yearly fatalities in lower-and-middle-income nations take place outside of hospitals and due to inadequate death registration systems, this leaves no cause of death (COD) information available \cite{b3}. Verbal autopsy (VA) technology was developed to address the requirement for COD information for researchers and legislators. Trained surveyors interview a close relative involved in taking care of the deceased about events surrounding their death and physicians and more recently  automated algorithms later code the surveys for a COD \cite{Danso}.

For a variety of CODs, VAs have shown to be $98\%$ specific \cite{b8} and while some studies show the textual part of the report to be of very little use in COD classification \cite{King}, other works have show that the information in the narrative text helps in making accurate diagnoses \cite{Manaka, b18}.

\subsection{Transfer Learning}
For many real life machine learning tasks, it can be difficult to collect a lot of data when taking on a new task. This is especially true in domains such as health where there are data privacy policies for patients in place. Datasets in these domains are also unique to individual institutions and as such it has grown to be a challenge to obtain satisfactory model performances using a small amount of training data.

\citet{Baxter} proposed the transfer learning technique to tackle this problem. This technique allows for knowledge from languages, tasks, or domains with lots of data to be transferred to areas with less data. Deep learning techniques have played a significant role in developing SOTA transfer learning techniques across numerous areas of applications because of their proved accuracies. Deep learning techniques are data driven and these advances have not shown equal development in the health domain. Those that are emerging in the clinical domain such as ClinicalBERT \cite{Huang}, the publicly available Clinical BERT Embeddings \cite{Alsentzer}, and biomedical and scientific literature like BioBERT \cite{Lee}, BioELMo \cite{probing}, SciBERT \cite{Beltagy} depend on data that is transcribed by medical professionals and is still unique to institutions to some extent.

We define transfer learning using a task and a domain where the latter ${D}$, is composed of a feature space $\chi$ and a marginal probability distribution $P(X)$ over the feature space with $X = \{x_i , ..., x_n\} \in \chi$.

A task ${T}$ on the other hand is composed of a label space ${Y}$, a prior distribution $P(Y)$ and a conditional probability distribution $P(Y|X)$ for a training pair of  $x_i \in X$ and $y_i \in {Y}$.

Given a source domain ${D_S}$, a source task ${T_S}$, a target domain ${D_T}$ and a target task ${T_T}$, transfer learning can happen when ${D_S} \neq {D_T}$ or ${T_S} \neq {T_T}$ and it learns the target conditional probability distribution $P_T (Y_T | X_T)$ in ${D_T}$ with the information gained from ${D_S}$ and ${T_S}$.

For text classification, $\chi$ is the space of sentence representations, $x_i$ the $i^{th}$ term vector corresponding to a sentence and $X$ is the sample of sentences used for training. This study focuses on cross-domain learning, a setting where the source domain is different from the and target domains differ ${D_S} \neq {D_T}$.

\subsubsection{Domain Adaptation}
The marginal probability distributions of the source ,$P_S$ and of the target $P_T$ domains can differ, ${P_S(XS)} \neq {P_T(XT)}$, a setting which is called domain adaptation.

Although domain adaptation is normally investigated in a single source domain, there is a unique case of domain adaptation where data from numerous sources is available for training, a paradigm known as multi-source domain adaptation. A number of works have been done in this setting including training a single model from combined data from multiple sources \cite{Aue}. Other techniques involve training separate models for each source domain and combining them by ensemble techniques with self training \cite{LiZong}, using a linear combination of the base models \cite{Mansour} and multi-tasking and linear combination \cite{WuHuang}. Neural Network-based models are the more recent ones and include attention based models \cite{KimStratos, SuYan}.

\section{Methods}
The experimental set up of the framework applied on a verbal autopsy (VA) dataset is presented in this section. The experiment is divided into three parts; part one focuses on the selection of the best vocabulary set for the ELMo \cite{Peters} language model. The second part compares the strategies of handling class imbalances in text classification and the third part focuses on cause of death (COD) classification of binary VA features, narrative text features and a hybrid of binary and narrative features settings of a verbal autopsy report.

\subsection{Algorithms}
BERT \cite{Devlin}, BERT Experts-PubMed \cite{tensorflow2015}, ELMo \cite{Peters2}, BioELMo \cite{Jin} and a feedforward neural network classifier.\\
ELMo and BERT embeddings were computed on Google Colaboratory and the pretrained models used were from Tensorflow Hub\footnote{https://www.tensorflow.org/hub}.

\subsection{Dataset}
The verbal autopsy (VA) dataset is from the MRC/Wits Rural Public Health and Health Transitions Research Unit (Agincourt), in South Africa \cite{Agincourt}, ethics clearance number:M110138. It is a unit that supports investigations into causes and impacts of diseases to social transitions and populations. The data was collected from 1992 to 2015 and consists of 8698 VAs. The VA records were reviewed for features suggestive of uncontrollable hyperglycemia by a clinician with paediatric training and expertise in type-1 diabetes management in high-income and low-and-middle-income countries. 3708 cases had symptoms of uncontrollable hyperglycemia and 77 were identified as positive and 7755 negative cases of death by uncontrollable hyperglycemia. Apart from the closed ended questions the data also has \enquote{free text} describing circumstances leading up to to death.

\begin {table}[htb!]
\caption{Verbal Autopsy Binary Features} \label{tab:VAfeature} 
\centering
\resizebox{\textwidth}{!}{%
\begin{tabular}{cccccccccc}
 \toprule
 \textbf{female} & \textbf{tuber} & \textbf{diabetes} & \textbf{men-con} & \textbf{cough} & \textbf{ch-cough} & \textbf{diarr} & \textbf{exc-urine} & \textbf{exc-drink} & \textbf{diagnosis}\\ 
 \bottomrule\toprule
  0 & 1 & 0 & 0 & 1 &	1 &	0 &	0 &	1 &	0\\
  1 & 0 & 0 & 0 & 1 &	1 &	1 &	0 &	0 &	0\\
  1 & 1 & 0 & 0 &	1 &	1 &	0 &	0 &	1 &	1\\
  1 & 0 & 0 & 0 &	0 &	0 &	0 &	0 &	0 &	0\\
  0 & 0 & 0 & 1 &	0 &	0 &	0 &	0 &	1 &	1\\
  1 &	1 &	0 &	1 &	1 &	1 &	1 &	0 &	0 &	0\\
 \bottomrule
\end{tabular}}
\end {table}

\begin{table}[htb!]
\centering
\caption{A Sample of Verbal Autopsy Narrative} \label{Verbal Narrative}
\resizebox{\columnwidth}{!}{%
  \begin{tabular}{l} 
  \toprule
  \textbf{Narrative}\\ 
  \bottomrule\toprule
  The deceased started by having painful abdomen. The following day she was taken to the clinic. The \\ nurse didn't say what was wrong. Treatment was given but nothing change. After few days. She was \\ feeling cold. She had difficult in breathing and she stop talking and walking. Where she was taken to \\ the health center. Where oxygen was given and referred to the hospital by an ambulance as an out \\ patient. The doctors said it was poison. Water drip, oxygen and treatment was given but no change. \\ She died the same day at the hospital while the doctors where helping her.\\
  \midrule
  \textbf{Diagnosis:} 0\\
  \toprule
\end{tabular}}
\end{table}

\begin{figure}[htb!]
    \centering
    \includegraphics[width=1.0\textwidth]{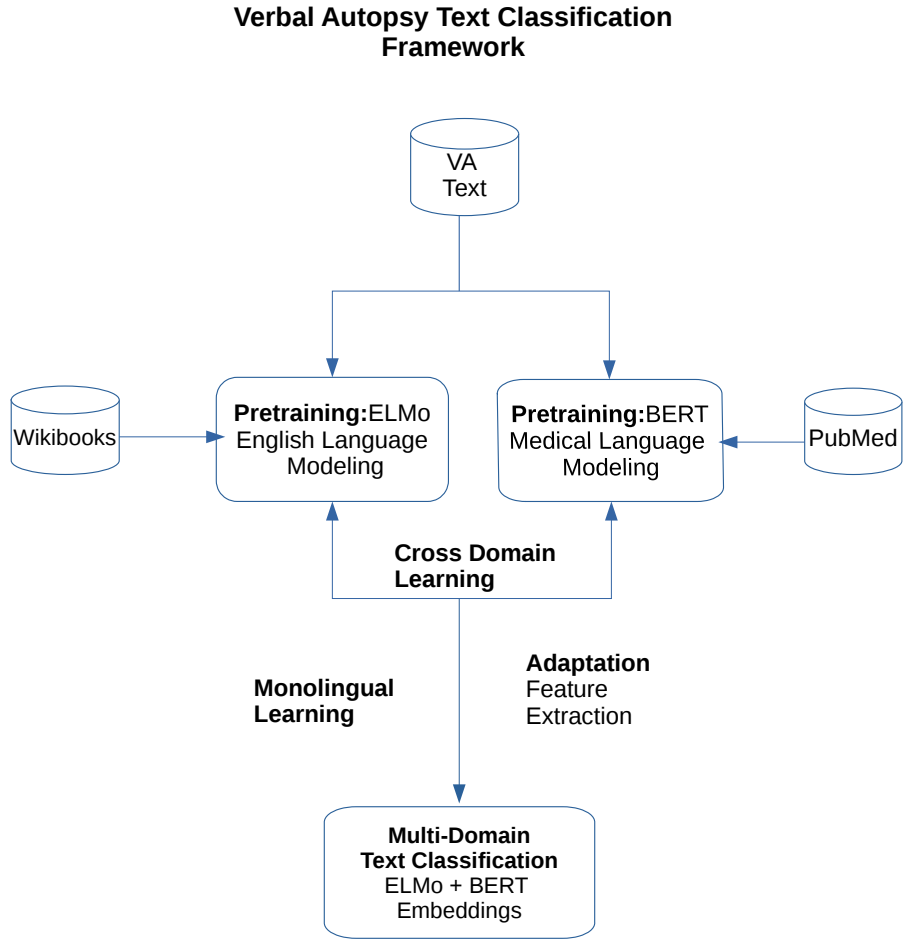}
    \caption{Text Classification Framework}
    \label{Figure.1}
\end{figure}

Binary features were identified as symptoms for which there was a  yes or no response, and these replies were transformed into 1's and 0's. These include the symptoms indicated in Table \ref{tab:VAfeature} along with excessive thirst, urination, mental disorientation, and others. Text features were assigned to the narratives outlining the deceased's signs and circumstances leading up to death; an example of one of these elements is shown in Table \ref{Verbal Narrative}.

We trained three ELMo models from scratch to evaluate the effectiveness of features taken from the English, medical, and public health domains. One vocabulary was built using the English Wikipedia and monolingual news crawl data from WMT 2008–2012, another the VA corpus, and the third one from PubMed abstracts. The datasets were preprocessed by text lowercasing and punctuation removal. The training data was then randomly divided into numerous training files, each including tokenized text with one sentence per line. We cloned the tensorflow implementation of ELMo from github repository \footnote{https://github.com/allenai/bilm-tf} and kept the same hyperparameters of the original ELMo and BioELMo models and trained the three ELMo models. We lastly computed the perplexity of the biLMs on test data and for inclusion in downstream task text classification, all layers of ELMo were collapsed into a single vector of size 1024.

We created a simple feature extraction model by creating a preprocessing model, a pretrained model from a family of BERT Experts models from Tensorflow Hub pretrained in different domains, one of which is the original BERT model trained on Wikibooks and the other which is pretrained on PubMed abstracts. For this study we will term the latter BERT-PubMed.

A map with three key values was created by the BERT model:pooled output which represents each input sequence as a whole, i.e. the embedding for all VA data, sequence output which represents each input token in context, i.e. the contextual embedding for every token in the VA corpus, and encoder outputs which represent intermediate activations in the transformer blocks. For extracted BERT embeddings we used the 768-element pooled output array.  

The ELMo emebeddings with the lowest perplexity scores of the three were combined with the BERT embeddings, yielding a vector size of 1792, and the two BERT versions were evaluated and compared on the text classification task. A multi-layered feed-forward neural network was utilized, with 5-fold cross validation incorporated in the process of training. Fig \ref{Figure.1} gives the multi-domain framework for text classification.

\subsection{Class Imbalance}

\textbf{SMOTE-Tomek Links}\\
A method called SMOTE-Tomek Links, which integrates the synthetic minority oversampling technique (SMOTE) \cite{SMOTE} and the Tomek Links (T-Links) \cite{Thai} undersampling, was employed to handle the dataset imbalance.

SMOTE creates new minority-class instances by combining previously existing minority-class examples along the border connecting all of their k-nearest neighbors while the Tomek Links undersampling technique identifies those sets of data points that are close yet fall into different classes \cite{Thai}. If $a$ and $b$ are instances of classes $A$ and $B$, respectively, then $a$ and $b$ are referred to as Tomek Links points if for the distance between them $d(a, b)$, $d(a, b) < d(a, c)$ or $d(a, b) < d(b, c)$ for another point $c$.

\noindent \textbf{Cost-sensitive classification}\\ 
Cost-sensitive learning addresses the class imbalance by changing the model's cost function so that misclassifications of training samples from the minority class are given more weight and hence are more costly.

For a single prediction $x_i$, the weighted cross entropy (CE) loss for class $j$, is given by

\begin{equation}
CE = - \frac{1}{N}{\sum_{i}  \alpha_{i} \sum_{j \in \{0,1\}} y_{ij} log p_{ij}}
\end{equation}

where $x_i$ belongs to a set of training instances $X$, associated with a label $y_i$ and $y_i$  $\epsilon$ $\{0, 1\}$,  and the predicted probabilities of the classes is $p_i$, where $p_i$ $\epsilon$  $[0,1]$. $\alpha_{i} \epsilon [0,1]$ is set by the inverse class frequency and it's value is equal to 1 when the loss function is not weighted.

We examined the two procedures for dealing with class imbalances and compared them to when no sampling was done. Following \cite{madabushi} we increased the weight of incorrectly labeling a VA case by changing the cost function of our model's fully connected layer during the training by multiplying each example's loss by a factor. The computed class weights ratio used is $0.50494305 : 51.07608696$. 

For data resampling we coupled the undersampling technique of Tomek Link with the oversampling method of SMOTE (SMOTE-Tomek Links). Each fold was sampled, and the classifier was trained on the training folds before being verified on the remaining folds. Classifier evaluation was done using recall, precision, F1-score, and area under the receiver operating characteristic curve (AUC-ROC). We tested the model with binary features, narrative text features, and a hybrid of binary and narrative text features. 

\section{Results and Discussion}

\begin{table}[htb!]
\centering
\caption{ELMo Language Models Evaluation:Perplexity}\label{tab2}
\begin{tabular}{llrrr}
\toprule
\textbf{Technique} & 
\textbf{Vocabulary} & 
\textbf{Tokens} & 
\textbf{\makecell[br]{Train\\Perplexity}} & 
\textbf{\makecell[br]{Test\\Perplexity}}\\
\midrule
\textbf{ELMo} & English Wikipedia & 5.5B & 43.23 & 37.32\\
\textbf{ELMo} & Verbal Autopsy & 982 495 & 71.55 & 50.01\\
\textbf{BioELMo} & PubMed Abstracts & 2.46B & 47.44 & 33.01\\
\bottomrule
\end{tabular}
\end {table}

The ELMo model trained on a Wikibooks and Book Corpus vocabulary achieved better perplexity scores on both the training and evaluation sets compared to the ELMo models pretrained on PubMed abstracts and a verbal autopsy (VA) vocabulary with the former performing better than the latter, as depicted in Table \ref{tab2} above. We believe that this is a result of the different sizes of the vocabulary sets derived from the three datasets and the fact that PubMed abstracts and the VA corpus contain numerous mentions of words in the English wikipedia and books where more linguistic knowledge including spellings, syntax and grammar are learned.

\begin {table}[htb!]
\centering
\caption{Class Imbalance Strategies on ELMo, BERT and Combined ELMo and BERT Embeddings on VA Text Features Setting} 
\label{Vectors}
\resizebox{\textwidth}{!}{%
  \begin{tabular}{ll|rrrrr} 
  \toprule
  \textbf{Model} & \textbf{Sampling} & \textbf{Recall} & \textbf{Precision} & \textbf{F1-Score} & \textbf{AUC-ROC} & \textbf{Accuracy} \\ 
  \bottomrule\toprule
  \multirow{3}{*}{\textbf{ELMo}}
  &\textbf{No Resampling} & 0 & 0 & 0 & 0.5 & 0.9898 \\
  &\textbf{SMOTE-Tomek Links} & 0.5000 & 0.0500 & 0.1005 & 0.7064 & 0.9086 \\
  &\textbf{Weighted Cross Entropy} & 0.5500 & 0.0159 & 0.0309 & 0.5999 & 0.7468 \\\midrule
  \multirow{3}{*}{\textbf{BERT}}
  &\textbf{No Resampling} & 0 & 0 & 0 & 0.5 & 0.9898 \\
  &\textbf{SMOTE-Tomek Links} & 0.6500 & 0.1044 & 0.1799 & 0.7667 & 0.8811 \\ 
  &\textbf{Weighted Cross Entropy} & 0.7000 & 0.1194 & 0.2040 & 0.7435 & 0.8351 \\\midrule
  \multirow{3}{*}{\textbf{BERT-PubMed}}
  &\textbf{No Resampling} & 0 & 0 & 0 & 0.5 & 0.9898 \\
  &\textbf{SMOTE-Tomek Links} & 0.6911 & 0.1135 & 0.1950 & 0.7966 & 0.8697 \\ 
  &\textbf{Weighted Cross Entropy} & 0.8133 & 0.1219 & 0.2120 & 0.7794 & 0.8800 \\\midrule
  \multirow{3}{*}{\makecell{\textbf{ELMo}\\\textbf{+BERT}}}
  &\textbf{No Resampling} & 0 & 0 & 0 & 0.5 & 0.9898 \\
  &\textbf{SMOTE-Tomek Links} & 0.7300 & 0.2000 & 0.3139 & 0.5724 & 0.9862 \\ 
  &\textbf{Weighted Cross Entropy} & 0.8455 & 0.2308 & 0.3626 & 0.7130 & 0.7187 \\\midrule
  \multirow{3}{*}{\makecell{\textbf{ELMo}\\\textbf{+BERT-PubMed}}}
  &\textbf{No Resampling} & 0 & 0 & 0 & 0.5 & 0.9898 \\
  &\textbf{SMOTE-Tomek Links} & 0.7654 & 0.2300 & 0.3537 & 0.6678 & 0.9782 \\ 
  &\textbf{Weighted Cross Entropy} & 0.8755 & 0.3146 & 0.4629 & 0.8413 & 0.8032 \\\midrule
  \bottomrule
  \end{tabular}
  }
\end {table}

\begin{figure}[htb!]
    \centering
    \subfloat{{\includegraphics[width=8cm, height=7cm]{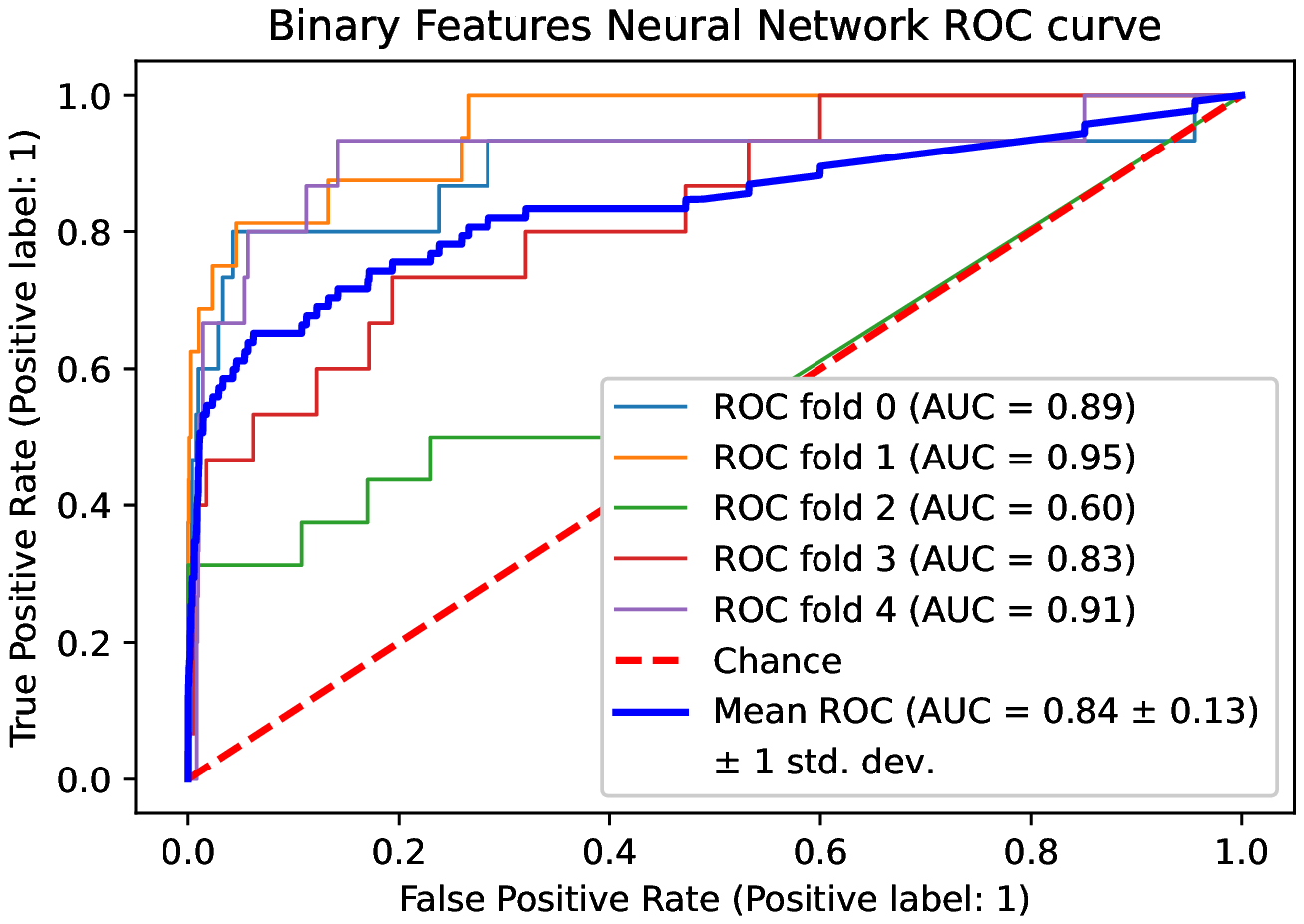}}}
    \caption{ROC curves and AUC-ROC of the Neural Network Classifier on Binary Features of a VA Report.}
    \label{Figure.2}
\end{figure}

\begin{figure}[htb!]
    \centering
    \subfloat{{\includegraphics[width=8cm, height=7cm]{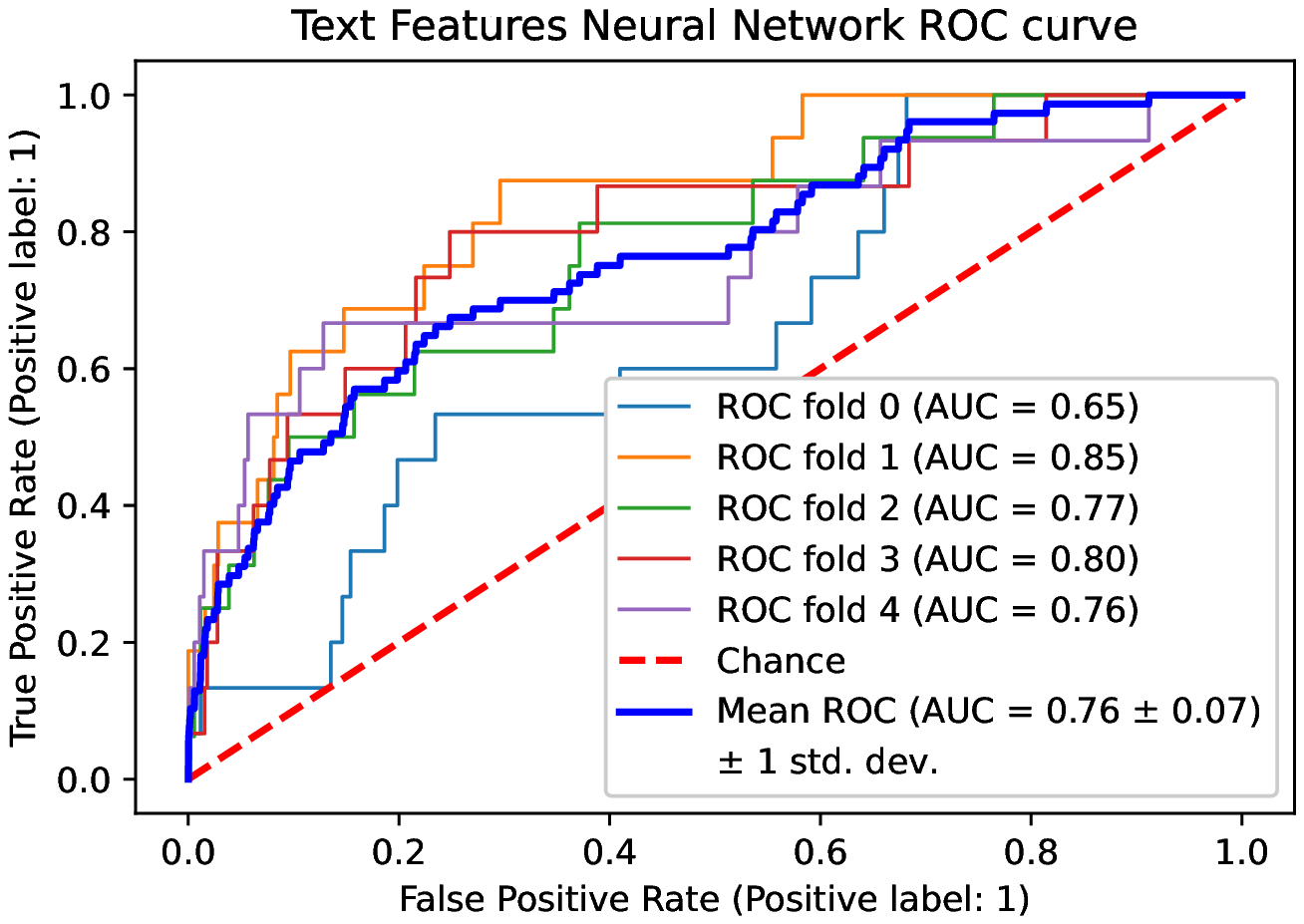}}}
    \caption{ROC curves and AUC-ROC of Neural Network classifier on Text Features of a VA Report.}
    \label{Figure.3}
\end{figure}

\begin{figure}[htb!]
    \centering
    \subfloat{{\includegraphics[width=8cm, height=7cm]{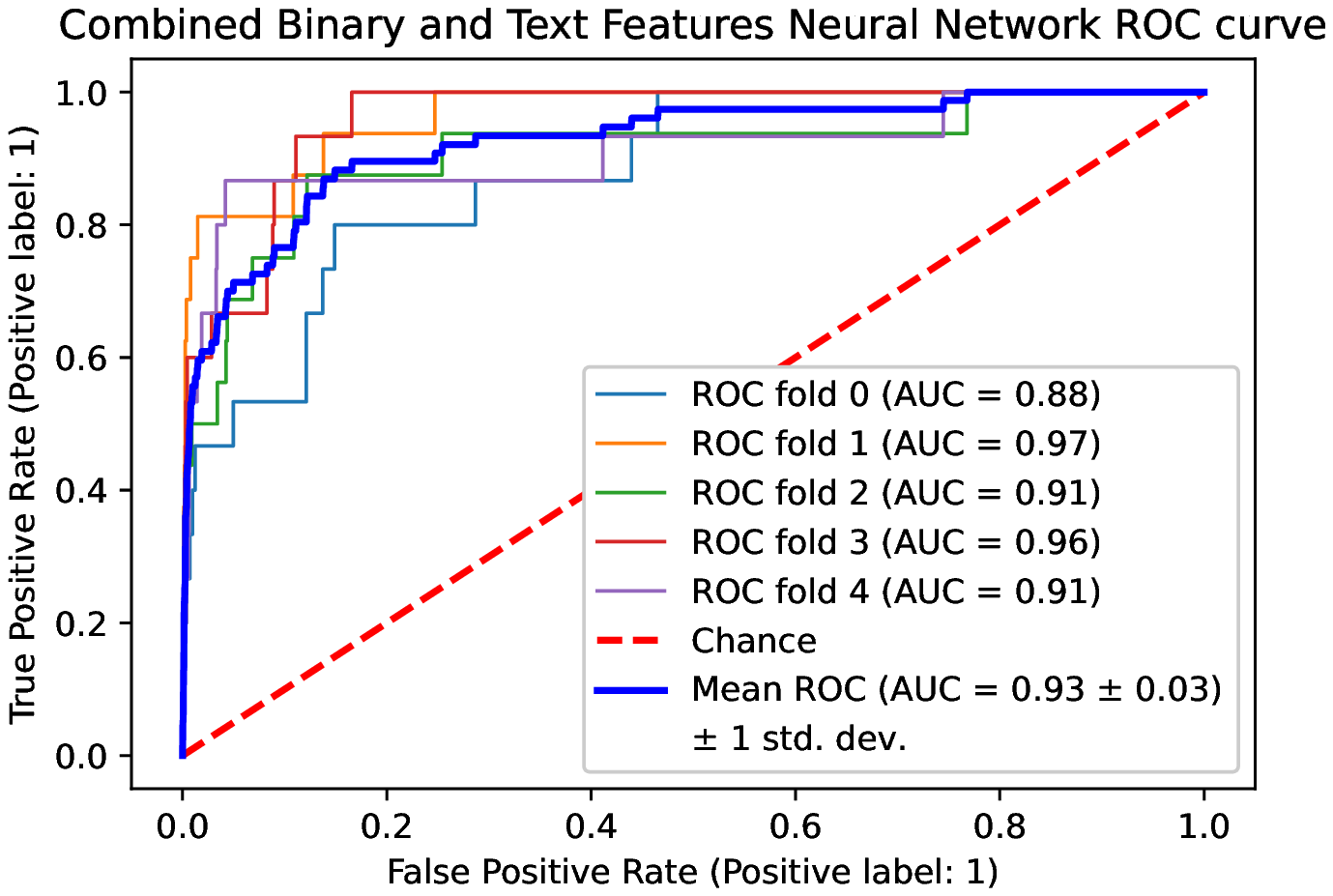}}}
    \caption{ROC curves and AUC-ROC of Neural Network classifier on the Combined Binary and Text Features of a VA Report.}
    \label{Figure.4}
\end{figure}

In general, the weighted binary cross entropy loss function produced better results across all metrics compared to the SMOTETomek sampling on the text classification task, as shown in Table \ref{Vectors}, and both are better than when no sampling is done at all. The performance scores achieved by weighted cross entropy loss function are however consistent with \citet{madabushi} who demonstrated that while BERT is capable of handling imbalanced datasets with no additional data augmentation, when the training and evaluation sets differ, as they do with VA reports, it struggles to generalize effectively.

\citet{madabushi} further show through an analysis on BERT with data augmentation and one without that data modifying techniques such as resampling and data augmentation techniques like synonym replacements do not lead to substantial improvements when using pretrained models like BERT. This is also consistent with the work of \citet{wei}, who tested the same methods for improving the text classification task.

Across all embedding comparisons in Table \ref{Vectors}, the neural network classifier achieved high recall scores and rather low precision scores. However with recall scores of around 0.8466 - 0.8755, we are convinced that character information in combination with word domain information can improve classification of cause of death (COD). It is also evident from these results that BERT pretrained on PubMed abstracts gives better embeddings well suited for COD classification than those from BERT pretrained on Wikibooks. These results are in line with \citet{probing} who show that word-level information from the medical domain is able to capture biomedical relations like symptom interactions and across different diseases, information that is crucial for COD classification.

The receiver operating characteristic (ROC) curve plots of the neural network classifier across the three VA features settings are given by Fig. \ref{Figure.2}, Fig. \ref{Figure.3} and Fig. \ref{Figure.4}.The ROC curves in each of the three settings all ascended toward the top left, indicating that the models successfully predicted both the cases. The combined features setting has the highest AUC-ROC score (93\%) in comparison to the narrative text and binary features separately, further proving the importance of text features in COD classification. 

Our results are consistent with those of \citet{Manaka}, who compared four machine learning algorithms in the binary VA features, textual VA features and a hybrid of textual and binary VA features. \citet{Manaka} attested that when the narrative text features are used in combination with the binary features, the background and depth of the events relating to death are enhanced, achieving an AUC-ROC score of 97\% on a neural network classifier. 

\subsubsection{Limitations of the Study}
This study is limited to the target task of text classification. More investigation can be done on the tasks of relation extraction and named entity recognition (NER) which can be considered important tasks in NLP in the health domain. Additionally, only English and clinical domains are used in this study, since this classification framework has significant impact on results, more experimentation could reveal whether or not similar behaviour occurs for other subsets of the English domain. It would also be interesting to investigate the same architecture with different character and word embedding models.

\section{Conclusion}\label{sec13}
We have experimentally shown that a multi-source domain adaptation can improve the cause of death classification task for verbal autopsy reports. Further, we have shown that this is possible in a setting where one language is a subset of another with the character-based language model used in the English domain and the word-based language model on the subset clinical domain. In our upcoming work, we will examine how well this architecture performs in other English subdomains, such as finance, and we will look into a variety of character-level and word-level embedding techniques.\

\section{Acknowledgments}
We are grateful to MRC/Wits-Agincourt for providing the verbal autopsy reports and guidance into understanding the dataset. Thokozile Manaka is supported by the United Nation's Organization of Women in Science for The Developing World (OWSD).

\bibliographystyle{unsrtnat}
\bibliography{sample}

\end{document}